\documentclass[10pt,twocolumn,letterpaper]{article}

\usepackage{wacv}              % To produce the CAMERA-READY version
\usepackage{graphicx}
\usepackage{amsmath}
\usepackage{amssymb}
\usepackage{booktabs}
\usepackage{multirow}
\usepackage{caption}
\usepackage{enumitem}
\usepackage{algorithm}
\usepackage{algpseudocode}
\usepackage{mathrsfs}
\usepackage[pagebackref,breaklinks,colorlinks]{hyperref}

\usepackage[capitalize]{cleveref}
\crefname{section}{Sec.}{Secs.}
\Crefname{section}{Section}{Sections}
\Crefname{table}{Table}{Tables}
\crefname{table}{Tab.}{Tabs.}

\def\wacvPaperID{1705} % *** Enter the WACV Paper ID here
\def\confName{WACV}
\def\confYear{2024}

\begin{document}

\title{Sign Language Production with Latent Motion Transformer}

% \author{Pan Xie   Peng Taiying\\
% Beihang University\\
% % \and
% % Yao Du\\
% % Beihang University\\
% % 37 Xueyuan Road, Haidian District, Beijing, P.R. China\\
% % {\tt\small duyaoo@buaa.edu.cn}
% \and
% {\tt\small zhangqipeng@buaa.edu.cn}
% }
\author{Pan Xie\quad Taiying Peng\thanks{Co-first author}\quad Yao Du
    \quad Qipeng Zhang  \\
    Beihang University, Beijing, China \\
    {\tt\small \{panxie, taiyi, duyaoo, zhangqipeng\}@buaa.edu.cn}
}

\maketitle

%%%%%%%%% ABSTRACT
\begin{abstract}
   
Sign Language Production (SLP) is the tough task of turning sign language into sign videos. The main goal of SLP is to create these videos using a sign gloss. In this research, we've developed a new method to make high-quality sign videos without using human poses as a middle step. Our model works in two main parts: first, it learns from a generator and the video's hidden features, and next, it uses another model to understand the order of these hidden features. To make this method even better for sign videos, we make several significant improvements. (i) In the first stage, we take an improved 3D VQ-GAN to learn downsampled latent representations. (ii) In the second stage, we introduce sequence-to-sequence attention to better leverage conditional information. (iii) The separated two-stage training discards the realistic visual semantic of the latent codes in the second stage. To endow the latent sequences semantic information, we extend the token-level autoregressive latent codes learning with perceptual loss and reconstruction loss for the prior model with visual perception. Compared with previous state-of-the-art approaches, our model performs consistently better on two word-level sign language datasets, i.e., WLASL and NMFs-CSL.
\end{abstract}

%%%%%%%%% BODY TEXT
\section{Introduction}
\label{sec:intro}

Sign language is the primary communication manner of the deaf community. As a visual language, it contains various hand gestures, movements, facial expressions, and complex grammatical structures. Therefore, it is difficult to communicate between sign and non-sign language speakers. To reduce the barrier, many studies~\cite{Koller2020WeaklySL,Cui2017RecurrentCN,Pu2018DilatedCN,Cheng2020FullyCN,Huang2018VideobasedSL,zhou2020spatial,camgoz2020multi,Camgz2018NeuralSL,Li2020TSPNetHF,9528010} have been dedicated to translating sign language into text and spoken language (SLR and SLT). More recent research~\cite{saunders2020adversarial,saunders2020progressive,saunders2021continuous,Zelinka_2020_WACV,stoll2020text2sign,Duarte_2021_CVPR} looks at doing the opposite: creating sign language videos from text (SLP). While most older SLP models focused on generating sign poses, Duarte~\textit{et al.}\cite{Duarte_2021_CVPR} and Stoll\textit{et al.}\cite{stoll2020text2sign} created sign videos in two steps: translating words to sign poses and then making videos from those signs. However, tools like OpenPose~\cite{8765346}, used in these processes, sometimes struggle with fast hand movements.

In this work, we explore a novel SLP method to bypass the human pose estimator and directly generate sign videos. In the area of video generation, one strand of work towards the variations of GANs~\cite{lee2018stochastic,mathieu2015deep,clark2019adversarial,vondrick2016generating,luc2020transformation}. Other strands of work propose variational autoencoders (VAEs)~\cite{denton2018stochastic,xue2016visual,babaeizadeh2017stochastic}, vector quantized VAE (VQ-VAE)~\cite{oord2017neural,razavi2019generating}, autoregressive models~\cite{weissenborn2019scaling}, and flows~\cite{kumar2019videoflow}. Moreover, a recent strand of work is latent autoregressive method, such as LVT~\cite{rakhimov2020latent}, VideoGPT~\cite{walker2021predicting}, and PV-VQVAE~\cite{yan2021videogpt}. They proposed to generate the video within two stages: first learn an image or a video generator using VQ-VAE, and then learn the prior of the latent codes with a causal Transformer (\textit{e.g.},~GPT~\cite{Radford2018ImprovingLU}) or convolution networks (\textit{e.g.},~PixelCNN~\cite{OordKEKVG16}). All these generative methods have their trade-offs in various aspects, \textit{e.g.}, sampling speed, sample quality, optimized stability, calculation requirements, and ease of evaluation. 
\begin{figure*}[t]
\centering
\includegraphics[width=0.9\linewidth]{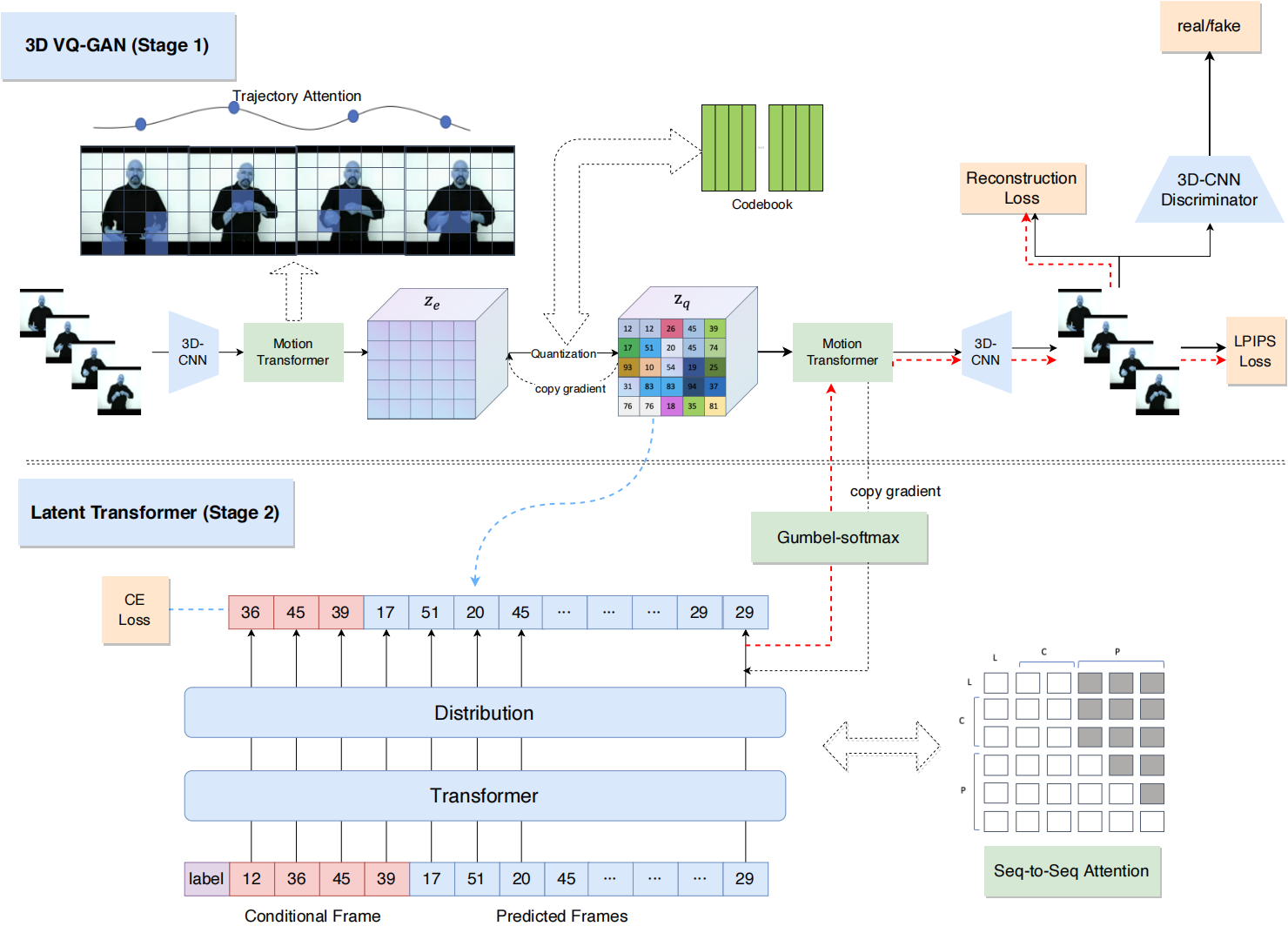}
\caption{The overview of our Latent Motion Transformer (LMT) with two-stage training strategy. In the first stage, the video VQ-GAN model encodes the sign videos into discrete codes, and learn the latent space with reconstruction loss, perceptual loss, and a discriminator. In the second stage training on the flattened codes, the conditional autoregressive model predicts the future codes with the cross-entropy loss, perceptual loss and reconstruction loss. Note that the decoder parameters is fixed in the second stage.}
\label{architecture}
\vspace{-0.0cm}
\end{figure*}

Our SLP model is also built upon the latent autoregressive method. We make this choice for several reasons: (i) Compared with the adversarial models, the two stages of the latent autoregressive model are optimized by the reconstruction and likelihood, which are more stable and easier to evaluate. This advantage allows us to focus on complex video modeling. (ii) Compared with the autoregressive models in the pixel space, performing autoregressive modeling in the downsampled latent space is much more efficient in training and sampling. 
% Standing on the shoulder of the previous latent autoregressive methods, 
To this end, we propose a novel \textbf{L}atent \textbf{M}otion \textbf{T}ransformer (LMT) for the word-level sign language production (see Figure \ref{architecture}). To better understand and produce high-quality sign videos, we make several modifications to existing methods. 

In the first stage training of reconstruction: \textbf{(i)} We leverage the VQ-GAN~\cite{esser2020taming}, a brilliant merger of VQ-VAE and GANs. Modifying its application, we transition it into a 3D-dimension for video reconstructions. This change involves the employment of 3D convolution networks to facilitate the transformation of pixel-space videos into a compact latent space. Such a downsampling technique is not arbitrary; it serves the essential purpose of optimizing computational load, especially for the subsequent latent codes prior learning phase. (ii) Deep within this reconstruction realm, we recognize the indispensability of the Transformer architecture in efficiently modeling the latent visual nuances~\cite{esser2020taming}. Many predecessors, like VideoGPT, have opted for axial attention~\cite{ho2019axial} to navigate the spatial-temporal latent space. Our divergence is towards the motion Transformer embedded with trajectory attention~\cite{patrick2021keeping}. This variant presents a superior capability in discerning the temporal dynamics innate in videos, particularly evident in the graceful choreography of hand movements in sign videos. Not to be overlooked, this methodology achieves an admirable optimization, reducing complexity to \(O(N)\). 

In the second stage of prior learning: \textbf{(iii)} We substitute the casual self-attention, embracing instead the sentence-to-sentence attention, which in our assessment, holds promise in effectively harnessing conditional frames and labels.

Furthermore, delving into historical approaches, \textbf{(iv)} there's a notable preference for the GPT-like model in prior works~\cite{rakhimov2020latent,walker2021predicting,yan2021videogpt,esser2020taming}. This method is inclined to unearth the latent discrete codes using a token-level cross-entropy loss function. However, such a bifurcated methodology appears to fall short in capturing the intricate visual semantics, risking subpar generation outcomes. Contrarily, our strategy integrates both the patch-level perceptual loss~\cite{zhang2018unreasonable} and the reconstruction loss. This dual-pronged approach seeks to endow the prior model with richer visual semantics. Executing this involves culling discrete codes from the Transformer model's predicted distribution. Despite the potential challenges (given its non-differentiable nature), we mitigate this through the Gumbel-softmax strategy~\cite{jang2016categorical}. Subsequent to this, the decoded predicted codes reveal the reconstructed videos, paving the way for a side-by-side comparison with their original versions, enabling computation of the perceptual and reconstruction losses.

Overall, our contributions are highlighted below:
\begin{itemize}[leftmargin=*]
\item[1.] We focus on generating realistic human-centered sign videos without using human pose sequences as the intermediate step. Furthermore, our innovative approach utilizes a 3D VQ-GAN with a motion transformer for latent sign video understanding, setting it apart from traditional axial attention and vanilla spatial-temporal attention methods.
\item[2.] Our method introduces sentence-to-sentence attention for prior learning on the flattened latent codes. This allows us to better harness conditional information and accurately predict coherent future frames. Additionally, our training method with patch-level perceptual loss and reconstruction loss enriches the prior model with visual perception, leading to a more potent model.
\item[3.] We conduct experiments on two public sign language datasets: WLASL~\cite{li2020word} and NMFs-CSL~\cite{hu2021global}. Compared with previous state-of-the-art video generation methods, our proposed LMT model achieves significant improvements with $-24.47$ and $-6.70$ FVD scores on two datasets, respectively.
\end{itemize}
%-------------------------------------------------------------------------
\section{Related Work}
% Our work is based on the two-stage approach which first learning an encoding of the data, and then learn a probabilistic model of the latent codes from the encoding.

\noindent\textbf{VQ-VAE and VQ-GAN.}
VQ-VAE~\cite{oord2017neural} is a specific auto-encoder model which consists of three modules: an encoder, a codebook, and a decoder. Given an image $x{\in} \mathbb{R}^{H\times W\times C}$, the encoder $E(x)$ first compresses the high dimensional data into low-dimension features $z_e{\in} \mathbb{R}^{h\times w\times n_z}$. Different from reconstructing on the compressed features directly, it performs a nearest neighbors method to quantize the compressed feature $z_e$ to the discrete representations $\hat z_q{\in} \mathbb{R}^{h\times w\times n_z}$, where the discrete vectors are maintained by the codebook $Z{=}\{e_i{\in} R^{n_z}\}^{K}_{i=1}$. This quantization process can be formulated as:
\begin{equation}
\begin{aligned}
{z_q}_{i,j} = e_{q(x_{i,j})},\quad  q(x_{i,j})=\mathop{argmin}\limits_{k\in K}\lVert {z_e}_{i,j} - e_k \rVert,
\end{aligned}
\label{eqn:equation1}
\end{equation}
where $q(\cdot)$ is the quantization method for calculating the corresponding index of the codebook embedding $Z$. 

After quantization, the decoder $D$ reconstruct the data from the discrete representations $z_q$. The whole model is optimized by minimizing the reconstruction loss and the gap between the $z_e$ and $z_q$ as the following object:
% \begin{equation}
% \begin{aligned}
% \mathcal{L}_{vqvae}(x) =& \begin{matrix}\underbrace{\lVert x-D(z_q)\rVert_2^2}\\ \mathcal{L}_{recon} \end{matrix} + \begin{matrix}\underbrace{\lVert sg[z_e] - z_q\rVert_2^2}\\ \mathcal{L}_{codebook} \end{matrix} \\
% &+ \beta \begin{matrix}\underbrace{\lVert sg[z_q] - z_e\rVert}\\ \mathcal{L}_{commit} \end{matrix}
% \end{aligned}
% \label{eqn:equation2}
% \end{equation}
\begin{equation}
\begin{aligned}
\mathcal{L}_{vqvae}(x) =& \underbrace{\lVert x-D(z_q)\rVert_2^2}_{\mathcal{L}_{recon}} + \underbrace{\lVert sg[z_e] - z_q\rVert_2^2}_{\mathcal{L}_{codebook}} \\
&+ \beta \underbrace{\lVert sg[z_q] - z_e\rVert}_{\mathcal{L}_{commit}},
\end{aligned}
\label{eqn:equation2}
\end{equation}
where $sg$ is the stop-gradient operator. $\mathcal{L}_{recon}$ is a reconstruction loss to encourage the model to learn meaningful representations for reconstructing the data. $\mathcal{L}_{codebook}$ is to bring the codebook vectors closer to the encoder output $z_e$. $\mathcal{L}_{commit}$ is a commit loss weighted by $\beta$ to prevent the encoder output from fluctuating between different code vectors. 

VQ-GAN~\cite{esser2020taming} is an improved model based on the VQ-VAE. To learn a perceptually rich codebook, it combines the reconstruction loss and a perceptual loss $\mathcal{L}_{lpips}$~\cite{zhang2018unreasonable}. Moreover, it introduces a patch-based discriminator $D$ to discriminate the real and reconstructed images. The adversarial training procedure is optimized by the following loss:

\begin{equation}
\begin{aligned}
\mathcal{L}_{gan}(x) = [logD(x) + log(1-D(G(z_q)))].
\end{aligned}
\label{eqn:equation3}
\end{equation}
Therefore, the complete objective for the VQ-GAN model can be formulated as:
% \begin{equation}
% \begin{aligned}
% \mathop{argmin}\limits_{E,G,C}\mathop{max}\limits_{D}\mathbb{E}_{x\in p(x)}[(\mathcal{L}_{vqvae} + \mathcal{L}_{lpips}) + \lambda\mathcal{L}_{gan}]
% \end{aligned}
% \label{eqn:equation4}
% \end{equation}
\begin{equation}
\begin{aligned}
\arg\min\limits_{E,G,C} \max\limits_{D}\mathbb{E}_{x\in p(x)}[(\mathcal{L}_{vqvae} + \mathcal{L}_{lpips}) + \lambda\mathcal{L}_{gan}].
\end{aligned}
\label{eqn:equation4}
\end{equation}
where $\lambda{=} \dfrac{\nabla_{G_L} [L_{recon}]}{\nabla_{G_L} [L_{gan}] + \delta}$is an adaptive weight. $\nabla_{G_L}[\cdot]$ means the gradient of its input to the last layer of the decoder $G_L$. And $\delta{=}10^{-6}$ is used for numerical stability.

\noindent\textbf{Vision Transformer.}
The Transformer architecture~\cite{vaswani2017attention} has led the way in many language tasks. In vision, some have used Transformers for image classification~\cite{dosovitskiy2020image,touvron2021training}. But in video, it's newer territory. A few works~\cite{bertasius2021space,arnab2021vivit} show promise using spatial-temporal attention for video. The main hurdle? Transformers are computationally heavy due to their quadratic complexity. Some solutions like Longformer~\cite{beltagy2020longformer}, Linformer\cite{wang2020linformer}, and Nyströmformer~\cite{xiong2021nystr} are trying to cut this down. We're using trajectory attention~\cite{patrick2021keeping} for dynamic scenes, which is faster due to its linear complexity.

\noindent\textbf{Autoregressive Vision Prediction Models.}
PixelCNNs~\cite{OordKEKVG16} and Image-GPT~\cite{chen2020generative} are key models for generating images. Weissenborn \textit{et al.}~\cite{weissenborn2019scaling} used this idea for videos, breaking down the data distribution $P(x)$ into conditional probabilities for all pixels:

\begin{equation}
\begin{aligned}
P(x) = \prod_{i=0}^{n}p_{\theta} (x_i|x_{<i}),
\end{aligned}
\label{eqn:equation5}
\end{equation}
where $n$ is the full dimensionality of the data. $x_{<i}$ means the pixels before $x_i$ in the raster-scan ordering. In this paper, we jointly train the autoregressive model with additional perceptual loss and reconstruction loss which endow the model with visual semantic awareness.

\section{Latent Motion Transformer}
Figure~\ref{architecture} illustrates the overall architecture of our proposed LMT. In the following subsections, we will give detailed descriptions about the two-stage training, \textit{i.e.}, the video VQ-GAN for latent codes learning and the latent Transformer model for prior learning. 

\subsection{3D VQ-GAN}
In the first stage, we utilize the 3D VQ-GAN to learn a set of latent codes on the sign videos. The 3D VQ-GAN is composed of three module: a encoder, a codebook and a decoder. 

\noindent \textbf{Encoder.} The encoder consists of a series of 3D convolution networks and a motion transformer network. Given a video $x{\in} \mathbb{R}^{T\times H\times W\times C}$, the 3D convolution networks reduce the size of the video to $x'{\in} \mathbb{R}^{t\times h\times w\times n_z}$, where $t{=}T/2^m$, $h{=}H/2^m$, $w{=}W/2^m$, and $m$ is the number of downsample blocks. 

\begin{figure}[t]
\centering
\includegraphics[width=1\linewidth]{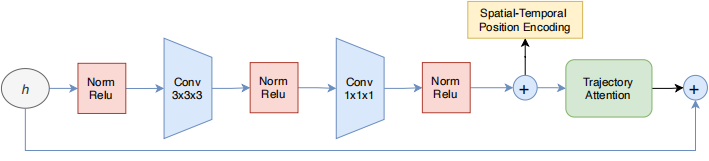}
\caption{The architecture of the attention residual block with trajectory attention.}
\label{attention} 
\vspace{-0.5cm}
\end{figure}

\noindent \textbf{Motion Transformer.} We use the Transformer network to understand the long-term interactions in videos, as depicted in Figure~\ref{attention}. Given its quadratic complexity, direct self-attention on codes from convolution networks is challenging. To solve this, we utilize trajectory attention~\cite{patrick2021keeping}, which focuses on hand movement trajectories. This method identifies a reference point in space and time and finds its trajectory by comparing with other points over time. An approximation technique further accelerates this attention mechanism, as detailed in Algorithm~\ref{algorithm1}.

\begin{algorithm}[h]
  \caption{Prototype-based attention.}
  \label{algorithm1}
  \begin{algorithmic}[1]
    \State P $\longleftarrow$ MostOrthogonalSubset $(Q,K,R)$;
    \State $\Omega_1 = \mathcal{S}(Q^TP/\sqrt{D})$;
    \State $\Omega_2 = \mathcal{S}(P^TK/\sqrt{D})$;
    \State $Y=\Omega_1(\Omega_2V)$;
    \label{code:fram:similarity}
  \end{algorithmic}
  
\end{algorithm}
\vspace{-0.5cm}

The attention matrix is approximated using intermediate prototypes, selected as the most orthogonal subset of the queries and keys, given a desired number of prototypes $R$. This algorithm reduces the computation dependency of the attention to a linear complexity $O(N)$. To focus on our main contributions, we omit the detailed architecture and refer readers to~\cite{patrick2021keeping} for reference.

\noindent \textbf{Codebook and Decoder.} The encoded low-dimension features are mapped into the discrete latent codes $z_q$ by calculating a nearest neighbor from the codebook embedding as shown in Eq.~\eqref{eqn:equation1}. Then we feed the corresponding embedding vector of $z_q$ to the decoder module. The decoder is composed of a motion Transformer and a series of 3D transposed convolution networks.

\noindent \textbf{Training.} The training objective of the video VQ-GAN is similar to the Eq.~\eqref{eqn:equation4}. The only difference is that our patch-level discriminator is composed of 3D convolutions instead of 2D convolutions used in image VQ-GAN~\cite{esser2020taming}.

\subsection{Latent Transformer}

In the second stage, we learn an autoregressive prior model over the flattened latent codes from the first stage. More precisely, the quantized encoding of a video is given by the encoder $E(x)$ and quantization $q(\cdot)$ as $z_q{=}q(E(x)){\in} \mathbb{R}^{t\times h\times w\times n_z}$. We flatten the spatial-temporal indices in the raster scan order, and the indices can be seen as a sequence $S {=} \{s_1, s_2, ...,s_i,.., s_{(t\times h\times w-1)}\}$, where $0 {\le} s_i {\le} |Z|-1$, $|Z|$ is the size of the codebook. Following the previous works, we generate the latent indices of videos in an autoregressive manner. Typically, the training objective of the autogressive model is expressed as a chain of conditional probabilities in a left-to-right manner:
\begin{equation}
\begin{aligned}
\mathcal{L}_{\text{ce}} = -\sum_{t=1}^{\small{t\times h\times w-1}}\log{p(s_i|s_{<i})},
\end{aligned}
\label{eqn:equation6}
\end{equation}
where $\mathcal{L}_{ce}$ is the cross-entropy loss function.

\begin{table*}[t]
\renewcommand\arraystretch{1.2}
\centering
\smallskip
\resizebox{0.85\textwidth}{!}{
\begin{tabular}{c cc cc cc cc}
\hline
\multirow{3}{*}{Models} & \multicolumn{4}{c}{\small{WLASL}} & \multicolumn{4}{c}{\small{NMFs-CSL}} \\ \cline{2-9}
 & \multicolumn{2}{c}{\small{Reconstruction}} & \multicolumn{2}{c}{\small{Predicting}} &
\multicolumn{2}{c}{\small{Reconstruction}} & \multicolumn{2}{c}{\small{Predicting}} \\
\hline
 & \small{R-FVD} $\downarrow$ & \small{LPIPS} $\downarrow$ & \small{FVD} $\downarrow$  & \small{SLR-Acc} $\uparrow$ & \small{R-FVD} $\downarrow$ & \small{LPIPS} $\downarrow$ & \small{FVD} $\downarrow$  & \small{SLR-Acc} $\uparrow$ \\
\hline
\multicolumn{9}{l}{\small{\textit{pose-sequence based method}}} \\
\hline
\small{Text2Sign~\cite{stoll2020text2sign}$^{*}$} & - & - & \small{342.16} & \small{0.360} & - & - & \small{261.02} & \small{0.468}  \\ 
\hline
\multicolumn{9}{l}{\small{\textit{GAN based method}}} \\
\hline
\small{TFGAN~\cite{ijcai2019-276}} & - & - & \small{357.87} & \small{0.348} & - & - & \small{268.22} &  \small{0.436}\\
\small{MoCoGAN-HD~\cite{TianRCO0MT21}} & - & - & \small{328.09} & \small{0.407} & - & - & \small{250.93} &  \small{0.517}\\
\hline
\multicolumn{9}{l}{\small{\textit{Latent Transformer method}}} \\
\hline
\small{LVT}~\cite{rakhimov2020latent} & \small{160.10} & \small{0.131} & \small{397.22} & \small{0.291} & \small{179.03} & \small{0.147} & \small{293.09} & \small{0.402}  \\
\small{VideoGPT}~\cite{yan2021videogpt} & \small{148.27} & \small{0.094} & \small{381.09} & \small{0.308} & \small{172.26} & \small{0.139}  & \small{281.27} & \small{0.419} \\
\small{\textbf{LMT} (Ours)} & \small{\textbf{120.77}} & \small{\textbf{0.048}} & \small{\textbf{301.62}} & \small{\textbf{0.430}} & \small{\textbf{144.59}} & \small{\textbf{0.055}} & \small{\textbf{244.23}} &  \small{\textbf{0.546}}\\
\hline
\end{tabular}}
\caption{The evaluation scores of all models on the two benchmark datasets. $\downarrow$ means the lower the better. $\uparrow$ means the higher the better. "${*}$" indicates the results obtained by our implementation.}
\label{comparison}
\vspace{-0.5cm}
\end{table*}

\noindent \textbf{Conditional Generation.} In our task, we generate the conditional sign videos given by a specifical label and a still human-center image. Specifically, we utilize a learnable embedding $L{\in} \mathbb{R}^{n_z\times n_{cls}}$ to embed the one-hot label to a dense vector $l{\in} \mathbb{R}^{n_z}$, where $n_{cls}$ is the number of sign language category. And we also encode the conditional still image into discrete latent codes $s_{c}$. As shown in Figure~\ref{architecture}, we introduce a sequence-to-sequence attention mechanism to better leverage the conditional information. In this attention mechanism, the latent codes of conditional frame and the label vector can see each other, but the future latent codes can only see previous codes before themselves with a causal mask. Then the conditional Transformer of our model can be formulated as:
\begin{equation}
\begin{aligned}
\mathcal{L}_{\text{ce}} = -\sum_{t=1}^{\small{t\times h\times w-1}}\log{p(s_i|s_{<i}, s_c, l)}.
\end{aligned}
\label{eqn:equation7}
\end{equation}

\subsection{Learning a Perceptually Prior Model} Many past studies mainly use the token-level cross-entropy loss when learning the prior model. However, this doesn't account for the visual meaning of the predicted codes, leading to a decline in generation quality. To address this, our approach first translates these predicted codes back into videos, then calculates the perceptual loss~\cite{zhang2018unreasonable} and the reconstruction loss. Due to the non-differentiability of predicting codes from the Transformer output, the Gumbel-softmax technique~\cite{jang2016categorical} is employed, using the straight-through estimator for gradient approximation~\cite{bengio2013estimating}. As illustrated by the red line in Figure~\ref{architecture}, after deducing the indices $\hat s$ and converting them to codebook entries $z_q(\hat s)$, we decode them into a video $\hat x{=}G(z_q(\hat s))$. The final step involves computing the perceptual and reconstruction losses by comparing the predicted video $\hat x$ to the actual video $x$. The complete objective of the latent Transformer is defined by:

\begin{equation}
\mathcal{L}_{\text{Transformer}} = \mathcal{L}_{ce} + \mathcal{L}_{lpips}(\hat x, x) + \mathcal{L}_{recon}(\hat x, x).
\label{eqn:equation8}
\end{equation}
To make a summary, equipping with our several improvements, we finally arrive at our proposed latent motion Transformer for sign language production.

\section{Experiments}
\subsection{Experimental Settings}
\noindent\textbf{Datasets.}
% \label{evaluate_metric}
We evaluate our proposed model on two public word-level sign language datasets: WLASL~\cite{Li2020WordlevelDS} and NMFs-CSL~\cite{Hu2020GloballocalEN}.
(i) WLASL is an American sign language dataset (ASL) that are collected from the web videos. It contains 2,000 words and 21,083 samples in total for training, validation, and test, respectively. Among them, the Top-100 and Top-300 most frequent words are released as WLASL100 and WLASL300, respectively.
(ii) NMFs-CSL is a Chinese sign language dataset (CSL) which are collected in the laboratory environment. It totally contains 1,067 words, 25,608 and 6,402 samples for training and testing, respectively.

\noindent \textbf{Evaluation Metrics.} Our evaluation consists of two stages, each with specific metrics:

For the initial reconstruction phase, we use: 
(i) LPIPS (Learned Perceptual Image Patch Similarity)~\cite{zhang2018unreasonable}, a frame-centric metric that gauges the perceptual likeness of two frames. 
(ii) R-FVD (reconstruction Fréchet Video Distance)~\cite{unterthiner2018towards}, a dynamic-oriented metric derived from comparing statistics of the I3D network trained on the Kinetics-400 dataset~\cite{carreira2017quo}.

In the second stage's assessment, we employ: 
(i) FVD to compare the resemblance between the sampled videos and the original ones. 
(ii) Due to the unsuitability of LPIPS for the sampling stage, we utilize an accuracy metric, SLR-Acc, for the conditional generation. This metric assesses if our videos align with the designated categories. We opt for the publicly available leading I3D model as our SLR. It showcases a Top-5 accuracy of $88.0\%$ on NMFs-CSL and $78.38\%$ on the WLASL300 dataset.

\noindent\textbf{Training Details.}
We standardize all image data to a range of $[-0.5, 0.5]$. Each video's resolution is set at $128{\times} 128$, with 16 frames sampled per sign word. Depending on the dataset, we adjust our sampling approach. For the web-sourced WLASL dataset, we apply uniform sampling. As NMFs-ASL is lab-gathered, with key frames predominantly in the middle, we sample 16 frames from the central 48 frames. During inference, we generate 16 frames for each sign word. In the second training stage, the initial static image serves as the conditional frame. For in-depth model specifics and hyperparameters, please refer to the Appendix.

In the first stage, we use the AdamW optimizer to optimize the generator and discriminator with learning rate as 3e-4. In the second stage, we use the AdamW optimizer to optimize the latent Transformer model. The learning rate is also set as 3e-4 with a cosine annealing schedule. We utilize up to 8 Nvidia GTX 3090 GPUs (24G) to achieve all the experiments.

\begin{figure*}[t]
\centering
\includegraphics[width=0.8\linewidth]{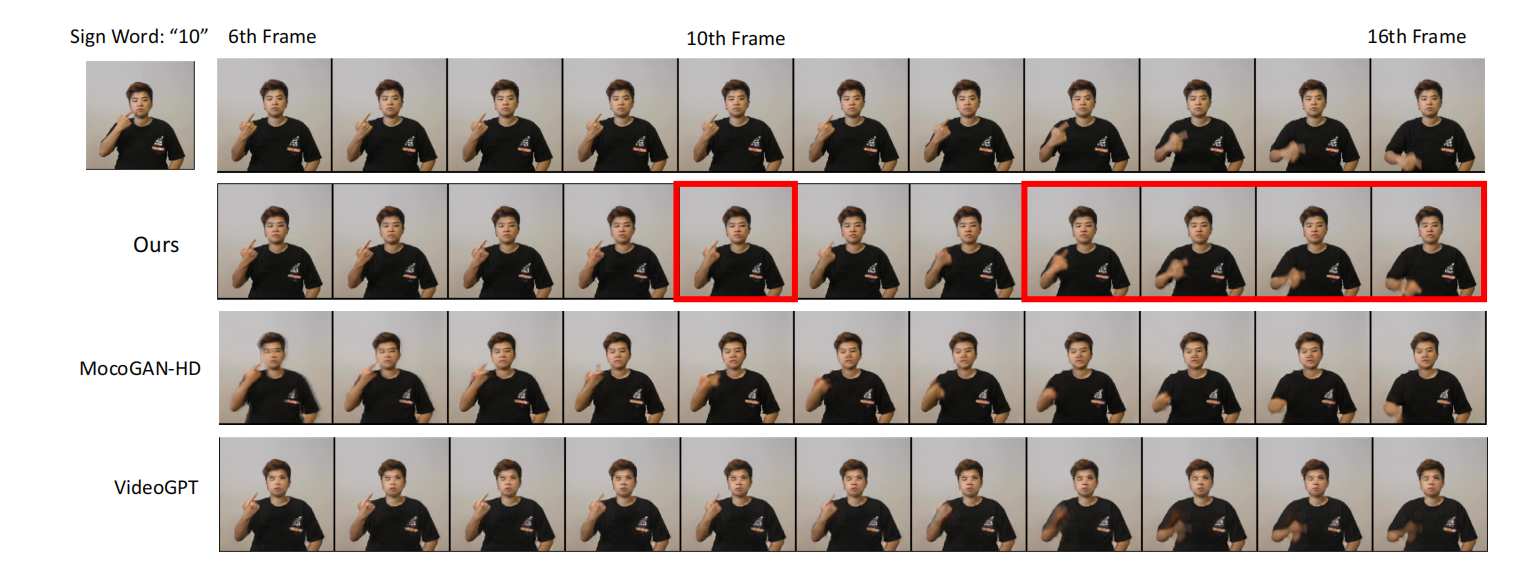}
\caption{The second stage predicted $128 {\times}128$ videos by the proposed LMT model on the NMFs-CSL dataset. The image in the left is the conditional frame, the sign word is the conditional category. Sign word "10" means the 10th category, since the corresponding word is not provided in the NMFs-CSL dataset. Note that the red boxes mean these frames with significant difference compared with the baselines, and our model get better generation quality.}
\label{nms_pred} 
\vspace{-0.0cm}
\end{figure*}

\begin{figure*}[t]
\centering
\includegraphics[width=0.8\textwidth]{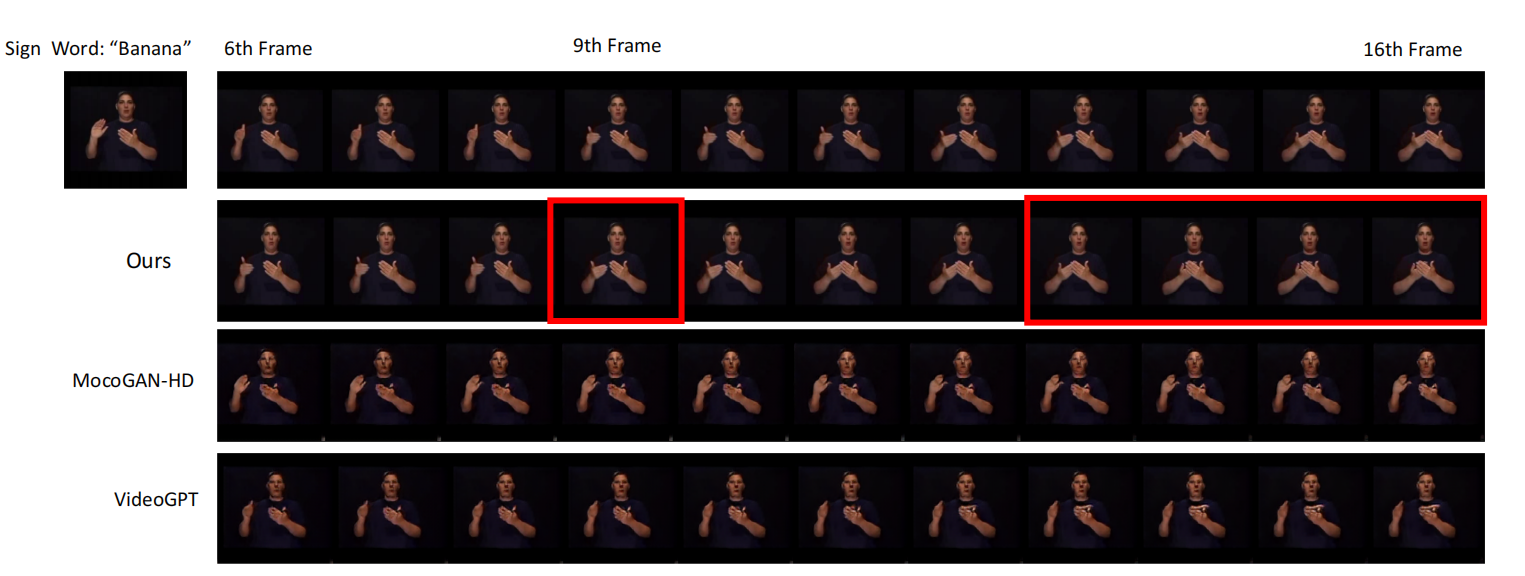}
\caption{The second stage predicted $128 {\times}128$ videos by the proposed LMT model on the WLASL dataset. The image in the left is the conditional frame, the sign word is the conditional category.}
\label{wlas_pred} 
\vspace{-0.0cm}
\end{figure*}

\subsection{Experimental Results}
In this section, we provide the quantitative comparison with existing methods for conditional word-level sign language production. We assess the performance of our model in terms of the mentioned evaluation metrics: R-FVD, LPIPS, FVD, and SLR-Acc. 

% Note that we are first attempting to predict sign video without sign pose sequence. Therefore, the baselines we choose to compare are the state-of-the-art latent autoregressive methods, \textit{i.e.}, VideoGPT~\cite{yan2021videogpt} and LMT~\cite{rakhimov2020latent}. In addition to the latent autoregressive methods, we also compare with the pose-sequence based method, \textit{i.e.}, Text2Sign~\cite{stoll2020text2sign}. And several GAN-based models, \textit{i.e.}, TFGAN~\cite{ijcai2019-276} and MoCoGAN-HD~\cite{TianRCO0MT21}.

\subsubsection{Baselines}

\noindent\textbf{Pose-sequence based method.} Text2Sign~\cite{stoll2020text2sign} converts spoken language into sign pose sequences using an NMT network and a motion graph, and then creates realistic sign videos from these poses. We reimplemented their method on our two word-level sign language datasets, adjusting the input to the NMT network to be a word rather than a sentence.

\noindent\textbf{GAN based method.} TFGAN~\cite{ijcai2019-276} is a conditional GAN with an effective multi-scale text-conditioning scheme based on discriminative convolutional filter generation. MoCoGAN-HD~\cite{TianRCO0MT21} is built on top of a pre-trained image generator~\cite{karras2019style}. It first learns the distribution of video frames as independent images, and then introduces a motion generator to discover the desired trajectory between the continuous latent codes, in which content and motion are disentangled.

\noindent\textbf{Latent Transformer method} VideoGPT~\cite{yan2021videogpt} and LVT~\cite{rakhimov2020latent} are both two-stage latent transformer methods which are similar to our proposed approach. They uses VQ-VAE to learn discrete latent representations of a raw video. And then a simple Transformer architecture is used to autoregressively model the discrete latents.

\subsubsection{Main Results.} Table~\ref{comparison} reveals our model's notable advancements over latent autoregressive methods like VideoGPT and LVT, evident in the R-FVD and FVD score improvements. Against GAN-based models, our results are superior by marked FVD score gains. Compared to pose-sequence methods, we show clear progress. Figures~\ref{nms_pred} and ~\ref{wlas_pred} showcase our model's capability in producing detailed future frames and consistent sign videos with accurate semantics. While our SLR-Acc outperforms others, all models still underperform in this metric, suggesting both the need for a more refined SLR model and challenges in generating precise sign videos, leaving scope for enhancement.

Furthermore, models perform differently on the WLASL and NMFs-CSL datasets. While they reconstruct better on WLASL, their predictive abilities are stronger on NMFs-CSL. This could be due to the shorter gesture duration and faster hand movements in NMFs-CSL. However, the lab-curated nature of NMFs-CSL, with central sign movements, aids in better understanding and forecasting.

\begin{figure*}[t]
\centering
\includegraphics[width=0.85\linewidth]{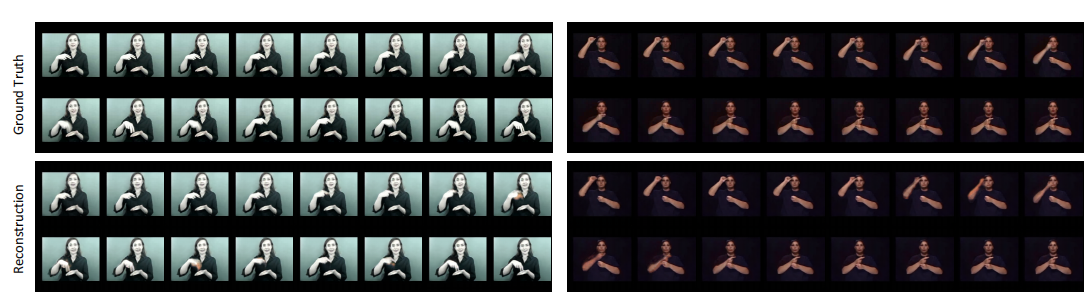}
\caption{The first stage reconstructed $128 {\times} 128$ videos by the video VQ-GAN with motion Transformer in the WLASL dataset. The two lines at the top are original video, two lines at the bottom are generated video.}
\label{wlasl_recon} 
\vspace{-0.0cm}
\end{figure*}

\begin{figure*}[t]
\centering
\includegraphics[width=0.85\linewidth]{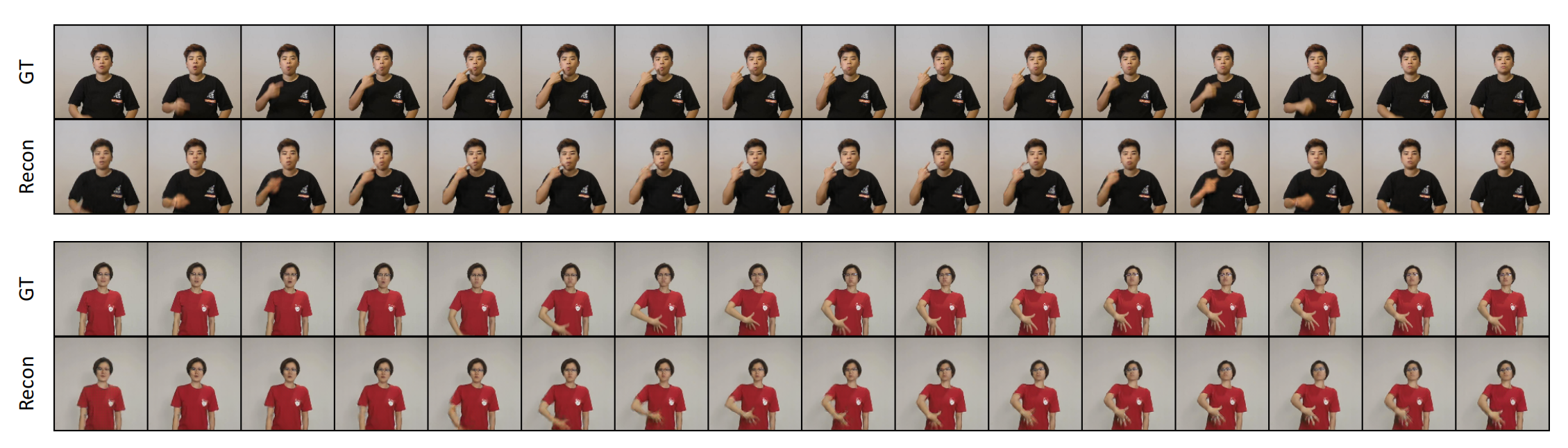}
\caption{The first stage reconstructed $128 {\times} 128$ videos by the video VQ-GAN with motion Transformer in the NMFs-CSL dataset. The two lines at the top are original video, two lines at the bottom are generated video.}
\label{nmscsl_recon} 
\vspace{-0.0cm}
\end{figure*}

\subsection{Ablation Study}
In this section, we present the ablation study that provides some intuitions as to why our approach works better than previous works.

\noindent\textbf{Do 3D VQ-GAN with Motion Transformer Generates High-Fidelity Sign Videos?}
In the first stage model, we make two improvements: replacing the VA-VAE model with VQ-GAN and incorporating the trajectory attention into the VQ-GAN architecture. To demonstrate the effectiveness of the modifications, we conduct comparable experiments on the WLASL dataset. As shown in Table~\ref{first_stage}, trajectory attention performs much better than axial attention which is used in VideoGPT~\cite{yan2021videogpt}. Moreover, by jointly training with perceptual loss and a patch-level discriminator, the VQ-GAN model can generate high-quality future frames than the VQ-VAE model. Figure~\ref{wlasl_recon} and Figure~\ref{nmscsl_recon} show some videos reconstructed by the video VQ-GAN with a motion Transformer on the two WLASL and NMFs-CSL datasets, respectively. 

\begin{table}[t]
\renewcommand\arraystretch{1.1}
\centering
\smallskip
\resizebox{0.4\textwidth}{!}{
\begin{tabular}{l c c}
\hline
\small{Model of First Stage} & \small{R-FVD} $\downarrow$ & \small{LPIPS} $\downarrow$ \\
\hline
\multicolumn{3}{l}{\small{\textit{VA-VAE}}} \\
\hline
\small{w/o Attention} & \small{162.74} & \small{0.107} \\
\small{Axial Attention} & \small{148.27} & \small{0.094}  \\
\small{Trajectory Attention} & \small{139.92} & \small{0.083} \\
\hline
\multicolumn{3}{l}{\small{\textit{VA-GAN (VQ-VAE + Perceptual Loss + GAN)}}} \\
\hline
\small{Axial Attention} & \small{138.97} & \small{0.069} \\
\small{Trajectory Attention (Ours)} & \small{\textbf{120.77}} & \small{\textbf{0.048}} \\
\hline
\end{tabular}}
\caption{The effect of modules in the first reconstruction stage on the WLASL dataset.}
\vspace{-0.0cm}
\label{first_stage}
\end{table}

\begin{table}[t]
\renewcommand\arraystretch{1.2}
\centering
\smallskip
\resizebox{0.48\textwidth}{!}{
\begin{tabular}{c c c c}
\hline
\small{Downsample Factor} & \small{Latent Size} & \small{R-FVD} $\downarrow$ & \small{LPIPS} $\downarrow$ \\
\hline
\small{(4,2,2)} & \small{$4\times 64\times 64$} & \small{129.17} & \small{0.120} \\
\hline
\small{(8,4,4)} & \small{$2\times 32\times 32$} & \small{137.13} & \small{0.061} \\
\small{(4,4,4)} & \small{$4\times 32\times 32$} & \small{120.77} & \small{0.048} \\
\small{(2,4,4)} & \small{$8\times 32\times 32$} & \small{\textbf{112.69}} & \textbf{\small{0.037}} \\
\hline
\small{(8,8,8)} & \small{$2\times 16\times 16$} & \small{147.81} & \small{0.075} \\
\small{(4,8,8)} & \small{$4\times 16\times 16$} & \small{140.26} & \small{0.068} \\
\small{(2,8,8)} & \small{$8\times 16\times 16$} & \small{132.94} & \small{0.053} \\
\hline
\small{(4,16,16)} & \small{$4\times 8\times 8$} & \small{167.33} & \small{0.107} \\
\hline
\end{tabular}}
\caption{The effect of spatial-temporal downsampling factor on the WLASL dataset.}
\label{downsample_factor}
\vspace{-0.5cm}
\end{table}

\noindent\textbf{Effect of the Spatial-Temporal Downsampling Factor.}
Choosing the right spatial-temporal downsampling factor for sign videos is crucial. In experiments on the WLASL dataset, as per Table~\ref{downsample_factor}, larger temporal latent sizes lead to improved reconstruction quality, and the trend is similar for spatial dimensions. However, the maximum spatial latent size of $64{\times} 64$ isn't optimal, suggesting patch-level Transformers outdo pixel-level ones.

The optimal spatial-temporal downsampling factor is $(2,4,4)$, equating to a latent size of $8{\times} 32{\times} 32$. Yet, bigger latent sizes increase computational demands in the subsequent Transformer model. Considering the computational complexity is roughly the square of the flattened codes' number, $8{\times} 32{\times} 32$ leads to a smaller model in the next training stage. Thus, a compromise is made, choosing $4{\times} 32{\times} 32$.

\noindent\textbf{Do Larger Size of Codebook Help?}
Unlike the vanilla VAE model, the VQ-VAE and VQ-GAN models use discrete latent codes that do not suffer from ``posterior collapse''. Similar to this collapse problem, using discrete latent space also faces the ``index collapse'', where only a few of the embedding vectors get trained due to a rich getting richer phenomenon. To explore this problem, we conduct experiments with different numbers of codebook vectors on the WLASL dataset. As shown in Table~\ref{size_of_codebook}, using 512 codes obtains a worse generation quality. Then increase the number of codes from 512 to 1024, the reconstruction and prediction performances are improved. When the code number reaches 2048, the performance is not affected. This phenomenon indicates that more codes increase the expression of the latent codebook, and using 1024 codes access a base threshold of the generation quality.

\begin{table}[t]
\renewcommand\arraystretch{1.1}
\centering
\smallskip
\resizebox{0.3\textwidth}{!}{
\begin{tabular}{c c c}
\hline
\small{Size of Codebook} & \small{R-FVD}$\downarrow$ & \small{FVD}$\downarrow$ \\
\hline
\small{2048} & \small{121.32} & \small{\textbf{300.25}} \\
\small{1024} & \small{\textbf{120.77}} & \small{301.62} \\
\small{512} & \small{136.65} & \small{324.11} \\
\hline
\end{tabular}}
\caption{The effect of the size of the codebook on the WLASL dataset.}
\label{size_of_codebook}
\vspace{-0.0cm}
\end{table}

\begin{table}[t]
\captionsetup{skip=2pt}
\renewcommand\arraystretch{1.2}
\centering
\smallskip
\resizebox{0.35\textwidth}{!}{
\begin{tabular}{l c c}
\hline
\small{Attention of Transformer} & \small{FVD}$\downarrow$ & \small{SLR-Acc}$\uparrow$ \\
\hline
\small{Casual Attention} & \small{319.36} & \small{0.416} \\
\small{Sent.-to-Sent. Attention (Ours)} & \small{\textbf{301.62}} & \small{\textbf{0.430}}  \\
\hline
\end{tabular}}
\caption{The effect of different attention mechanism over the flattened codes in the second stage on the WLASL dataset.}
\label{casual_attention}
\vspace{-0.5cm}
\end{table}

\noindent\textbf{Effect of Different Attentions over the Flattened Latent Codes.}
In the second stage of prior training, we replace the pure casual attention with the sentence-to-sentence attention to better leverage the conditional information. The attention mask mechanism is shown in Figure~\ref{architecture}, the latent codes of the conditional frame can see each other, thus learning a better representation of a single frame. In Table~\ref{casual_attention}, we compare LMT with different attention mechanism. We can see that our proposed attention brings benefits to our generation quality. 

\begin{figure}[t]
\centering
\includegraphics[width=0.75\linewidth]{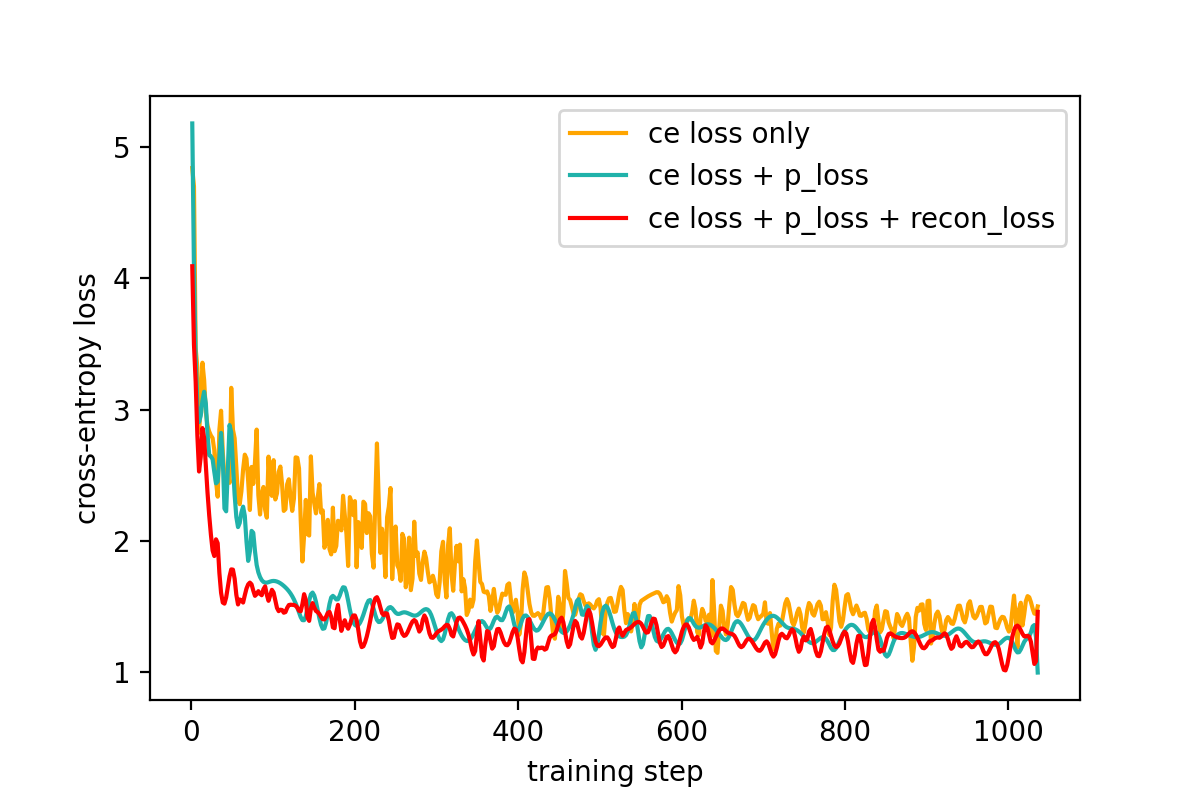}
\caption{The cross-entropy loss curves of training prior Transformer model incorporating with perceptual loss and reconstruction loss on the WLASL dataset.}
\label{ce_loss} 
\vspace{-0.0cm}
\end{figure}

\noindent\textbf{Do Perceptual Loss and Reconstruction Loss Help for Prior Model Learning?}
Traditional latent autoregressive transformer models are primarily optimized using token-level cross-entropy loss, focusing on the autoregressive nature of the codes to produce meaningful latent representations. This method, however, lacks a visual evaluation of the generated codes. As evidenced in Table~\ref{prior_loss}, training the model with both perceptual and reconstruction losses results in a substantial FVD improvement of $-58.58$. Figure~\ref{ce_loss} further illustrates the training loss curves. By integrating these two additional losses, the Transformer not only converges quicker but also yields a lower cross-entropy loss. This highlights that incorporating perceptual and reconstruction losses grants the model better visual understanding, leading to a more proficient prior model.

\begin{table}[t]
\captionsetup{skip=2pt}
\renewcommand\arraystretch{1.1}
\centering
\smallskip
\resizebox{0.35\textwidth}{!}{
\begin{tabular}{l c c}
\hline
\small{Traning Objective} & \small{FVD}$\downarrow$ & \small{SLR-Acc}$\uparrow$  \\
\hline
\small{Only CE Loss} & \small{360.20} & \small{0.335}  \\
\small{$\quad$+ Peceptual Loss} & \small{316.05}  & \small{0.417} \\
\small{$\quad$+ Reconstruction Loss (Ours)} & \textbf{\small{301.62}}  & \small{\textbf{0.430}} \\
\hline
\end{tabular}}
\caption{The effect of perceptual loss and the reconstruction loss on the prior training on the WLASL dataset. The third lines means training three losses together.}
\label{prior_loss}
\vspace{-0.5cm}
\end{table}

\subsection{Limitations}
Using a two-stage strategy, our model produces conditional sign videos, a more intricate approach than one-stage methods. While it creates coherent, high-quality frames, it has potential for better accuracy in generating videos for specific sign words.
\section{Conclusion}

In this work, we propose a novel  Latent Motion Transformer (LMT) model to realize the word-level sign language production without human pose sequences. We've developed a two-stage model inspired by recent latent autoregressive approaches to enhance sign video generation. First, we utilize the VQ-GAN for effective latent code learning and the motion Transformer to track hand movements. In the next stage, we integrate sentence-to-sentence attention for better conditional and categorical data use. Instead of using the cross-entropy loss function, we combine perceptual and reconstruction losses, resulting in enhanced generation quality. Tests on two sign language datasets confirm our model's superior performance against existing methods.

%-------------------------------------------------------------------------

%%%%%%%%% REFERENCES
{\small
\bibliographystyle{ieee_fullname}
\bibliography{reference}

\begin{thebibliography}{10}\itemsep=-1pt

\bibitem{arnab2021vivit}
Anurag Arnab, Mostafa Dehghani, Georg Heigold, Chen Sun, Mario Lu{\v{c}}i{\'c},
  and Cordelia Schmid.
\newblock Vivit: A video vision transformer.
\newblock {\em ArXiv preprint}, 2021.

\bibitem{babaeizadeh2017stochastic}
Mohammad Babaeizadeh, Chelsea Finn, Dumitru Erhan, Roy~H. Campbell, and Sergey
  Levine.
\newblock Stochastic variational video prediction.
\newblock In {\em 6th International Conference on Learning Representations,
  {ICLR} 2018, Vancouver, BC, Canada, April 30 - May 3, 2018, Conference Track
  Proceedings}. OpenReview.net, 2018.

\bibitem{ijcai2019-276}
Yogesh Balaji, Martin~Renqiang Min, Bing Bai, Rama Chellappa, and Hans~Peter
  Graf.
\newblock Conditional gan with discriminative filter generation for
  text-to-video synthesis.
\newblock In {\em Proceedings of the Twenty-Eighth International Joint
  Conference on Artificial Intelligence, {IJCAI-19}}, pages 1995--2001.
  International Joint Conferences on Artificial Intelligence Organization, 7
  2019.

\bibitem{beltagy2020longformer}
Iz Beltagy, Matthew~E Peters, and Arman Cohan.
\newblock Longformer: The long-document transformer.
\newblock {\em ArXiv preprint}, 2020.

\bibitem{bengio2013estimating}
Yoshua Bengio, Nicholas L{\'e}onard, and Aaron Courville.
\newblock Estimating or propagating gradients through stochastic neurons for
  conditional computation.
\newblock {\em ArXiv preprint}, 2013.

\bibitem{bertasius2021space}
Gedas Bertasius, Heng Wang, and Lorenzo Torresani.
\newblock Is space-time attention all you need for video understanding?
\newblock In {\em Proceedings of the 38th International Conference on Machine
  Learning, {ICML} 2021, 18-24 July 2021, Virtual Event}, Proceedings of
  Machine Learning Research, pages 813--824. {PMLR}, 2021.

\bibitem{Camgz2018NeuralSL}
Necati~Cihan Camg{\"{o}}z, Simon Hadfield, Oscar Koller, Hermann Ney, and
  Richard Bowden.
\newblock Neural sign language translation.
\newblock In {\em 2018 {IEEE} Conference on Computer Vision and Pattern
  Recognition, {CVPR} 2018, Salt Lake City, UT, USA, June 18-22, 2018}, pages
  7784--7793. {IEEE} Computer Society, 2018.

\bibitem{camgoz2020multi}
Necati~Cihan Camgoz, Oscar Koller, Simon Hadfield, and Richard Bowden.
\newblock Multi-channel transformers for multi-articulatory sign language
  translation.
\newblock In {\em European Conference on Computer Vision (ECCV)}, pages
  301--319. Springer, 2020.

\bibitem{8765346}
Z. Cao, G. Hidalgo, T. Simon, S. Wei, and Y. Sheikh.
\newblock Openpose: Realtime multi-person 2d pose estimation using part
  affinity fields.
\newblock {\em IEEE Transactions on Pattern Analysis and Machine Intelligence},
  pages 172--186, 2021.

\bibitem{carreira2017quo}
Jo{\~{a}}o Carreira and Andrew Zisserman.
\newblock Quo vadis, action recognition? a new model and the kinetics dataset.
\newblock In {\em 2017 {IEEE} Conference on Computer Vision and Pattern
  Recognition, {CVPR} 2017, Honolulu, HI, USA, July 21-26, 2017}, pages
  4724--4733. {IEEE} Computer Society, 2017.

\bibitem{chen2020generative}
Mark Chen, Alec Radford, Rewon Child, Jeffrey Wu, Heewoo Jun, David Luan, and
  Ilya Sutskever.
\newblock Generative pretraining from pixels.
\newblock In {\em Proceedings of the 37th International Conference on Machine
  Learning, {ICML} 2020, 13-18 July 2020, Virtual Event}, Proceedings of
  Machine Learning Research, pages 1691--1703. {PMLR}, 2020.

\bibitem{Cheng2020FullyCN}
Ka~Leong Cheng, Zhaoyang Yang, Qifeng Chen, and Yu-Wing Tai.
\newblock Fully convolutional networks for continuous sign language
  recognition.
\newblock In {\em European Conference on Computer Vision (ECCV)}, pages
  697--714. Springer, 2020.

\bibitem{clark2019adversarial}
Aidan Clark, Jeff Donahue, and Karen Simonyan.
\newblock Adversarial video generation on complex datasets.
\newblock {\em ArXiv preprint}, 2019.

\bibitem{Cui2017RecurrentCN}
Runpeng Cui, Hu Liu, and Changshui Zhang.
\newblock Recurrent convolutional neural networks for continuous sign language
  recognition by staged optimization.
\newblock In {\em 2017 {IEEE} Conference on Computer Vision and Pattern
  Recognition, {CVPR} 2017, Honolulu, HI, USA, July 21-26, 2017}, pages
  1610--1618. {IEEE} Computer Society, 2017.

\bibitem{denton2018stochastic}
Emily Denton and Rob Fergus.
\newblock Stochastic video generation with a learned prior.
\newblock In {\em Proceedings of the 35th International Conference on Machine
  Learning, {ICML} 2018, Stockholmsm{\"{a}}ssan, Stockholm, Sweden, July 10-15,
  2018}, Proceedings of Machine Learning Research, pages 1182--1191. {PMLR},
  2018.

\bibitem{dosovitskiy2020image}
Alexey Dosovitskiy, Lucas Beyer, Alexander Kolesnikov, Dirk Weissenborn,
  Xiaohua Zhai, Thomas Unterthiner, Mostafa Dehghani, Matthias Minderer, Georg
  Heigold, Sylvain Gelly, Jakob Uszkoreit, and Neil Houlsby.
\newblock An image is worth 16x16 words: Transformers for image recognition at
  scale.
\newblock In {\em 9th International Conference on Learning Representations,
  {ICLR} 2021, Virtual Event, Austria, May 3-7, 2021}. OpenReview.net, 2021.

\bibitem{Duarte_2021_CVPR}
Amanda Duarte, Shruti Palaskar, Lucas Ventura, Deepti Ghadiyaram, Kenneth
  DeHaan, Florian Metze, Jordi Torres, and Xavier Giro-i Nieto.
\newblock How2sign: A large-scale multimodal dataset for continuous american
  sign language.
\newblock In {\em Proceedings of the IEEE/CVF Conference on Computer Vision and
  Pattern Recognition (CVPR)}, pages 2735--2744, 2021.

\bibitem{esser2020taming}
Patrick Esser, Robin Rombach, and Björn Ommer.
\newblock Taming transformers for high-resolution image synthesis.
\newblock In {\em 2020 IEEE/CVF Conference on Computer Vision and Pattern
  Recognition (CVPR)}, 2021.

\bibitem{ho2019axial}
Jonathan Ho, Nal Kalchbrenner, Dirk Weissenborn, and Tim Salimans.
\newblock Axial attention in multidimensional transformers.
\newblock {\em ArXiv preprint}, 2019.

\bibitem{Hu2020GloballocalEN}
Hezhen Hu, Wengang Zhou, Junfu Pu, and H. Li.
\newblock Global-local enhancement network for nmfs-aware sign language
  recognition.
\newblock {\em ArXiv preprint}, 2020.

\bibitem{hu2021global}
Hezhen Hu, Wengang Zhou, Junfu Pu, and Houqiang Li.
\newblock Global-local enhancement network for nmf-aware sign language
  recognition.
\newblock {\em ACM transactions on multimedia computing, communications, and
  applications (TOMM)}, pages 1--19, 2021.

\bibitem{Huang2018VideobasedSL}
Jie Huang, Wengang Zhou, Qilin Zhang, Houqiang Li, and Weiping Li.
\newblock Video-based sign language recognition without temporal segmentation.
\newblock In {\em Proceedings of the Thirty-Second {AAAI} Conference on
  Artificial Intelligence, (AAAI-18), the 30th innovative Applications of
  Artificial Intelligence (IAAI-18), and the 8th {AAAI} Symposium on
  Educational Advances in Artificial Intelligence (EAAI-18), New Orleans,
  Louisiana, USA, February 2-7, 2018}, pages 2257--2264. {AAAI} Press, 2018.

\bibitem{jang2016categorical}
Eric Jang, Shixiang Gu, and Ben Poole.
\newblock Categorical reparameterization with gumbel-softmax.
\newblock In {\em 5th International Conference on Learning Representations,
  {ICLR} 2017, Toulon, France, April 24-26, 2017, Conference Track
  Proceedings}. OpenReview.net, 2017.

\bibitem{karras2019style}
Tero Karras, Samuli Laine, and Timo Aila.
\newblock A style-based generator architecture for generative adversarial
  networks.
\newblock In {\em Proceedings of the IEEE/CVF conference on computer vision and
  pattern recognition}, pages 4401--4410, 2019.

\bibitem{Koller2020WeaklySL}
Oscar Koller, N.~C. Camgoz, H. Ney, and R. Bowden.
\newblock Weakly supervised learning with multi-stream cnn-lstm-hmms to
  discover sequential parallelism in sign language videos.
\newblock {\em IEEE Transactions on Pattern Analysis and Machine Intelligence},
  pages 2306--2320, 2020.

\bibitem{kumar2019videoflow}
Manoj Kumar, Mohammad Babaeizadeh, Dumitru Erhan, Chelsea Finn, Sergey Levine,
  Laurent Dinh, and Durk Kingma.
\newblock Videoflow: {A} conditional flow-based model for stochastic video
  generation.
\newblock In {\em 8th International Conference on Learning Representations,
  {ICLR} 2020, Addis Ababa, Ethiopia, April 26-30, 2020}. OpenReview.net, 2020.

\bibitem{lee2018stochastic}
Alex~X Lee, Richard Zhang, Frederik Ebert, Pieter Abbeel, Chelsea Finn, and
  Sergey Levine.
\newblock Stochastic adversarial video prediction.
\newblock {\em ArXiv preprint}, 2018.

\bibitem{li2020word}
Dongxu Li, Cristian Rodriguez, Xin Yu, and Hongdong Li.
\newblock Word-level deep sign language recognition from video: A new
  large-scale dataset and methods comparison.
\newblock In {\em The IEEE Winter Conference on Applications of Computer
  Vision}, pages 1459--1469, 2020.

\bibitem{Li2020WordlevelDS}
Dongxu Li, Cristian Rodriguez-Opazo, X. Yu, and Hongdong Li.
\newblock Word-level deep sign language recognition from video: A new
  large-scale dataset and methods comparison.
\newblock {\em 2020 IEEE Winter Conference on Applications of Computer Vision
  (WACV)}, pages 1448--1458, 2020.

\bibitem{Li2020TSPNetHF}
Dongxu Li, Chenchen Xu, Xin Yu, Kaihao Zhang, Benjamin Swift, Hanna Suominen,
  and Hongdong Li.
\newblock Tspnet: Hierarchical feature learning via temporal semantic pyramid
  for sign language translation.
\newblock In {\em Advances in Neural Information Processing Systems 33: Annual
  Conference on Neural Information Processing Systems 2020, NeurIPS 2020,
  December 6-12, 2020, virtual}, 2020.

\bibitem{luc2020transformation}
Pauline Luc, Aidan Clark, Sander Dieleman, Diego de~Las Casas, Yotam Doron,
  Albin Cassirer, and Karen Simonyan.
\newblock Transformation-based adversarial video prediction on large-scale
  data.
\newblock {\em ArXiv preprint}, 2020.

\bibitem{mathieu2015deep}
Micha{\"{e}}l Mathieu, Camille Couprie, and Yann LeCun.
\newblock Deep multi-scale video prediction beyond mean square error.
\newblock In {\em 4th International Conference on Learning Representations,
  {ICLR} 2016, San Juan, Puerto Rico, May 2-4, 2016, Conference Track
  Proceedings}, 2016.

\bibitem{patrick2021keeping}
Mandela Patrick, Dylan Campbell, Yuki~M Asano, Ishan Misra~Florian Metze,
  Christoph Feichtenhofer, Andrea Vedaldi, Jo Henriques, et~al.
\newblock Keeping your eye on the ball: Trajectory attention in video
  transformers.
\newblock {\em ArXiv preprint}, 2021.

\bibitem{Pu2018DilatedCN}
Junfu Pu, Wengang Zhou, and Houqiang Li.
\newblock Dilated convolutional network with iterative optimization for
  continuous sign language recognition.
\newblock In {\em Proceedings of the Twenty-Seventh International Joint
  Conference on Artificial Intelligence, {IJCAI} 2018, July 13-19, 2018,
  Stockholm, Sweden}, pages 885--891. ijcai.org, 2018.

\bibitem{Radford2018ImprovingLU}
Alec Radford and Karthik Narasimhan.
\newblock Improving language understanding by generative pre-training.
\newblock 2018.

\bibitem{rakhimov2020latent}
Ruslan Rakhimov, Denis Volkhonskiy, Alexey Artemov, Denis Zorin, and Evgeny
  Burnaev.
\newblock Latent video transformer.
\newblock In {\em 16th International Joint Conference on Computer Vision,
  Imaging and Computer Graphics Theory and Applications, VISIGRAPP 2021}, pages
  101--112. SciTePress, 2021.

\bibitem{razavi2019generating}
Ali Razavi, A{\"{a}}ron van~den Oord, and Oriol Vinyals.
\newblock Generating diverse high-fidelity images with {VQ-VAE-2}.
\newblock In {\em Advances in Neural Information Processing Systems 32: Annual
  Conference on Neural Information Processing Systems 2019, NeurIPS 2019,
  December 8-14, 2019, Vancouver, BC, Canada}, pages 14837--14847, 2019.

\bibitem{saunders2020adversarial}
Ben Saunders, Richard Bowden, and Necati~Cihan Camg{\"{o}}z.
\newblock Adversarial training for multi-channel sign language production.
\newblock In {\em 31st British Machine Vision Conference 2020, {BMVC} 2020,
  Virtual Event, UK, September 7-10, 2020}. {BMVA} Press, 2020.

\bibitem{saunders2020progressive}
Ben Saunders, Necati~Cihan Camgoz, and Richard Bowden.
\newblock {Progressive Transformers for End-to-End Sign Language Production}.
\newblock In {\em Proceedings of the European Conference on Computer Vision
  (ECCV)}, 2020.

\bibitem{saunders2021continuous}
Ben Saunders, Necati~Cihan Camgoz, and Richard Bowden.
\newblock {Continuous 3D Multi-Channel Sign Language Production via Progressive
  Transformers and Mixture Density Networks}.
\newblock In {\em International Journal of Computer Vision (IJCV)}, 2021.

\bibitem{stoll2020text2sign}
Stephanie Stoll, Necati~Cihan Camgoz, Simon Hadfield, and Richard Bowden.
\newblock Text2sign: Towards sign language production using neural machine
  translation and generative adversarial networks.
\newblock {\em International Journal of Computer Vision (IJCV)}, pages
  891--908, 2020.

\bibitem{TianRCO0MT21}
Yu Tian, Jian Ren, Menglei Chai, Kyle Olszewski, Xi Peng, Dimitris~N. Metaxas,
  and Sergey Tulyakov.
\newblock A good image generator is what you need for high-resolution video
  synthesis.
\newblock In {\em 9th International Conference on Learning Representations,
  {ICLR} 2021, Virtual Event, Austria, May 3-7, 2021}, 2021.

\bibitem{touvron2021training}
Hugo Touvron, Matthieu Cord, Matthijs Douze, Francisco Massa, Alexandre
  Sablayrolles, and Herv{\'{e}} J{\'{e}}gou.
\newblock Training data-efficient image transformers {\&} distillation through
  attention.
\newblock In {\em Proceedings of the 38th International Conference on Machine
  Learning, {ICML} 2021, 18-24 July 2021, Virtual Event}, Proceedings of
  Machine Learning Research, pages 10347--10357. {PMLR}, 2021.

\bibitem{unterthiner2018towards}
Thomas Unterthiner, Sjoerd van Steenkiste, Karol Kurach, Raphael Marinier,
  Marcin Michalski, and Sylvain Gelly.
\newblock Towards accurate generative models of video: A new metric \&
  challenges.
\newblock {\em ArXiv preprint}, 2018.

\bibitem{OordKEKVG16}
A{\"{a}}ron van~den Oord, Nal Kalchbrenner, Lasse Espeholt, Koray Kavukcuoglu,
  Oriol Vinyals, and Alex Graves.
\newblock Conditional image generation with pixelcnn decoders.
\newblock In {\em Advances in Neural Information Processing Systems 29: Annual
  Conference on Neural Information Processing Systems 2016, December 5-10,
  2016, Barcelona, Spain}, pages 4790--4798, 2016.

\bibitem{oord2017neural}
A{\"{a}}ron van~den Oord, Oriol Vinyals, and Koray Kavukcuoglu.
\newblock Neural discrete representation learning.
\newblock In {\em Advances in Neural Information Processing Systems 30: Annual
  Conference on Neural Information Processing Systems 2017, December 4-9, 2017,
  Long Beach, CA, {USA}}, pages 6306--6315, 2017.

\bibitem{vaswani2017attention}
Ashish Vaswani, Noam Shazeer, Niki Parmar, Jakob Uszkoreit, Llion Jones,
  Aidan~N. Gomez, Lukasz Kaiser, and Illia Polosukhin.
\newblock Attention is all you need.
\newblock In {\em Advances in Neural Information Processing Systems 30: Annual
  Conference on Neural Information Processing Systems 2017, December 4-9, 2017,
  Long Beach, CA, {USA}}, pages 5998--6008, 2017.

\bibitem{vondrick2016generating}
Carl Vondrick, Hamed Pirsiavash, and Antonio Torralba.
\newblock Generating videos with scene dynamics.
\newblock In {\em Advances in Neural Information Processing Systems 29: Annual
  Conference on Neural Information Processing Systems 2016, December 5-10,
  2016, Barcelona, Spain}, pages 613--621, 2016.

\bibitem{walker2021predicting}
Jacob Walker, Ali Razavi, and A{\"a}ron van~den Oord.
\newblock Predicting video with vqvae.
\newblock {\em ArXiv preprint}, 2021.

\bibitem{wang2020linformer}
Sinong Wang, Belinda~Z Li, Madian Khabsa, Han Fang, and Hao Ma.
\newblock Linformer: Self-attention with linear complexity.
\newblock {\em ArXiv preprint}, 2020.

\bibitem{weissenborn2019scaling}
Dirk Weissenborn, Oscar T{\"{a}}ckstr{\"{o}}m, and Jakob Uszkoreit.
\newblock Scaling autoregressive video models.
\newblock In {\em 8th International Conference on Learning Representations,
  {ICLR} 2020, Addis Ababa, Ethiopia, April 26-30, 2020}. OpenReview.net, 2020.

\bibitem{9528010}
Pan Xie, Mengyi Zhao, and Xiaohui Hu.
\newblock Pisltrc: Position-informed sign language transformer with
  content-aware convolution.
\newblock {\em IEEE Transactions on Multimedia (TMM)}, pages 1--1, 2021.

\bibitem{xiong2021nystr}
Yunyang Xiong, Zhanpeng Zeng, Rudrasis Chakraborty, Mingxing Tan, Glenn Fung,
  Yin Li, and Vikas Singh.
\newblock Nyströmformer: A nyström-based algorithm for approximating
  self-attention.
\newblock {\em ArXiv preprint}, 2021.

\bibitem{xue2016visual}
Tianfan Xue, Jiajun Wu, Katherine~L. Bouman, and Bill Freeman.
\newblock Visual dynamics: Probabilistic future frame synthesis via cross
  convolutional networks.
\newblock In {\em Advances in Neural Information Processing Systems 29: Annual
  Conference on Neural Information Processing Systems 2016, December 5-10,
  2016, Barcelona, Spain}, pages 91--99, 2016.

\bibitem{yan2021videogpt}
Wilson Yan, Yunzhi Zhang, Pieter Abbeel, and Aravind Srinivas.
\newblock Videogpt: Video generation using vq-vae and transformers.
\newblock {\em ArXiv preprint}, 2021.

\bibitem{Zelinka_2020_WACV}
Jan Zelinka and Jakub Kanis.
\newblock Neural sign language synthesis: Words are our glosses.
\newblock In {\em Proceedings of the IEEE/CVF Winter Conference on Applications
  of Computer Vision (WACV)}, 2020.

\bibitem{zhang2018unreasonable}
Richard Zhang, Phillip Isola, Alexei~A. Efros, Eli Shechtman, and Oliver Wang.
\newblock The unreasonable effectiveness of deep features as a perceptual
  metric.
\newblock In {\em 2018 {IEEE} Conference on Computer Vision and Pattern
  Recognition, {CVPR} 2018, Salt Lake City, UT, USA, June 18-22, 2018}, pages
  586--595. {IEEE} Computer Society, 2018.

\bibitem{zhou2020spatial}
Hao Zhou, Wengang Zhou, Yun Zhou, and Houqiang Li.
\newblock Spatial-temporal multi-cue network for continuous sign language
  recognition.
\newblock In {\em The Thirty-Fourth {AAAI} Conference on Artificial
  Intelligence, {AAAI} 2020, The Thirty-Second Innovative Applications of
  Artificial Intelligence Conference, {IAAI} 2020, The Tenth {AAAI} Symposium
  on Educational Advances in Artificial Intelligence, {EAAI} 2020, New York,
  NY, USA, February 7-12, 2020}, pages 13009--13016. {AAAI} Press, 2020.

\end{thebibliography}
}

\end{document}